\documentclass{article}
\usepackage{spconf,amsmath,graphicx}
\usepackage{amsfonts}
\usepackage{algorithm}
\usepackage{algorithmic}

\usepackage{epsfig}
\usepackage{subfig}

\usepackage{epstopdf}
\usepackage{indentfirst}

\usepackage{color}
\usepackage{amscd}
\usepackage{colortbl}
\usepackage{xcolor}

\newcommand{\RR}{\mathbb{R}}
 \newcommand{\argmin}{\mathop{\mathrm{argmin}}}
\newcommand{\LL}{ \mathcal{L}}
\newcommand{\SC}{ \mathcal{S}}
\newcommand{\SP}{ \mathcal{P}}

\title{Inertial Proximal Deep Learning Alternating Minimization for Efficient  Neutral Network Training}
\name{Linbo Qiao$^{*}$\qquad Tao Sun$^{*}$\qquad Hengyue Pan \qquad Dongsheng Li$^{\dagger}$ 
\thanks{
$^{*}$The first two authors contributed equally.\newline
$~~~~~~~~~^{\dagger}$Dongsheng Li is the corresponding author.\newline
$~~~~~~~~$This work is sponsored in part by the National Key R\&D Program of China under Grant (2018YFB0204300)  and the National Natural Science Foundation of China under Grants (61806216, 61932001 and 61906200).}}
\address{College of Computer, National University of Defense Technology\\ \textit{qiao.linbo@nudt.edu.cn}, \textit{nudtsuntao@163.com},  \textit{\{hengyuepan, dsli\}@nudt.edu.cn}\\
}
\begin{document}

\maketitle
\begin{abstract}
In recent years, the Deep Learning Alternating Minimization (DLAM), which is actually  the alternating minimization applied to the penalty form of the deep neutral networks training, has been developed as an alternative algorithm to overcome several drawbacks of Stochastic Gradient Descent (SGD) algorithms.
This work develops an improved DLAM by the well-known inertial technique, namely iPDLAM, which predicts a point by linearization of current and last iterates.
To obtain further training speed, we apply a warm-up technique to the penalty parameter, that is, starting with a small initial one and increasing it in the iterations.
Numerical results on real-world datasets are reported to demonstrate the efficiency of our proposed algorithm.
\end{abstract}
\section{Introduction}

The deep neural network has achieved great success in computer vision and machine learning. Mathematically, training a $L$-layer neural network can be formulated as:
\begin{equation*}
\small
\min_{{\bf W}_1,{\bf W}_2,\ldots,{\bf W}_L} \{\LL({\bf y},\sigma_{L}({\bf W}_L ...\sigma_1({\bf W}_1{\bf a}_0)))+\sum_{l=1}^L R_l({\bf W}_l) \},
\end{equation*}
where ${\bf a}_0$ denotes the training sample and ${\bf y}$ denotes the labelled set, $R_l$ is the regularization in the $l$-th layer, and $\sigma_l$ is the $l$-th layer's activation.
The absence of $R_l$ yields the fully connected networks.
The main workhorse of the deep neutral network training  task is the SGD \cite{robbins1951stochastic} and its variants.
The core part of SGD for training neural networks is to compute the gradient with respect to ${\bf W}_l$, i.e., the   backpropagation \cite{rumelhart1988learning}.
For simplicity of presentation, we assume $\sigma_l$ is a $\RR\mapsto\RR$ mapping.
According to the chain rule, it follows:
$
\frac{\partial \LL}{\partial {\bf W}_1}=\frac{\partial \LL}{\partial \sigma_L}\frac{\partial \sigma_L}{\partial \sigma_{L-1}}\ldots\frac{\partial \sigma_2}{\partial \sigma_{1}}\frac{\partial \sigma_1}{\partial {\bf W}_1}.
$
If $|\frac{\partial \sigma_l}{\partial \sigma_{l-1}}|<1$, the gradient decays fast or even vanishes when $L$ is large.
This phenomenon, which has been mentioned in \cite{hochreiter2001gradient}, hurts the speed and performance of SGD in the deep layer case.
On the other hand, the convergence of stochastic training method is based on the Lipschitz continuous assumption on the gradient, which fails to hold for various applications.
To overcome these drawbacks, papers \cite{taylor2016training,zhang2016efficient,zhang2017convergent,wang2019admm} propose  gradient free methods  by the Alternating Direction Methods of Multipliers (ADMM) or Alternating Minimization.
The core idea of this  method is the decomposition of  the training task into a sequence of substeps which are just related to one-layer activations.
Due to that each substep can always find  its global minimizer, the gradient free method can achieve notable speedups \cite{taylor2016training,zhang2016efficient,zhang2017convergent}.
Another advantage of the gradient free is parallelism due to that ADMM and AM have natural conveniences to be implemented concurrently.
Besides the acceleration and parallelism, ADMM and ADM also enjoy  the advantage of mild theoretical guarantees to converge compared with SGD: The theory of SGD is heavily dependent on the global smoothness of  $\LL({\bf y},\sigma_{L}({\bf W}_L ...\sigma_1({\bf W}_1{\bf a}_0)))$ with respect to ${\bf W}_1,{\bf W}_2,\ldots,{\bf W}_L$, which usually fails to hold.
This paper studies the alternating minimization based method, namely Deep Learning Alternating Minimization (DLAM) and aims to develop an improved version of DLAM.

The training task can be reformulated as a nonlinear constrained optimization problem.
DLAM is actually the AM method applied to its penalty. In \cite{askari2018lifted}, the authors propose a new framework based on using  convex functions to appropriate a non-decreasing activation.
The AM method is the state-of-art solver for  the convex optimization problem.
In paper \cite{choromanska2018beyond}, the authors extend the DLAM  to the online learning with co-activation memory technique.
By representing the activation function as a  proximal operator form, \cite{li2019lifted} proposed a new penalty framework, which can also be minimized by AM.
With rewriting equivalent biconvex constraints to  activation functions, \cite{gu2018fenchel} proposes AM for Fenchel lifted networks.
With a higher dimensional space to lift the ReLU function, paper \cite{zhang2017convergent} develops a smooth multi-convex formulation with AM.
In \cite{carreira2014distributed}, the authors develop the AM methodology for the nested neural network.

The contribution of this paper can be concluded as follows:
1) A novel algorithm. We develop a new algorithm for deep neural networks training. In each iteration, the initial technique is employed to predict an auxiliary point with current and last iterate.
For acceleration,  we use a small penalty parameter in the beginning iterations and then  increase it to a larger one.
The DLAM can be regarded as a special case of this scheme.
2) Sound results. Various proximal operaters are widely used in statistical learning comunity \cite{combettes2011proximal}, which bring various applications problems potentially could be solved by the proposed algorithm in this work.
We present the numerical experiments to demonstrate the performance of our algorithm.
The convergence is verified, and comparisons with latest classical solvers are presented.

\section{Problem Formulation and Algorithms}

\noindent \textbf{Notation:} Given an integer  {\small$N>0$, $[N]:=\{1,2,\ldots,N\}$.}
Let $(a^k)_{k\geq 1}$ and $(b^k)_{k\geq 1}$ be positive, $a^k=o(b^k)$ means $\lim_{k}\frac{a^k}{b^k}=0$.
We denote that ${\bf z}^k:=({\bf z}_{1}^{k},\ldots,{\bf z}_{l}^k,\ldots,{\bf z}_{L}^{k})$ and $[{\bf z}^k;{\bf z}_l]:=({\bf z}_{1}^{k+1},\ldots,{\bf z}_{l-1}^{k+1},{\bf z}_{l},{\bf z}_{l+1}^{k},\ldots,{\bf z}_{L}^{k})$.
Similar notations are defined for ${\bf W}^k$ and ${\bf a}^k$.
$\LL(\cdot)$ represents the loss function and $\|\cdot\|$ is the L2 norm.
For a map $\sigma(\cdot):\RR^d\mapsto\RR^d$ and vectors ${\bf w}, {\bf v}\in\RR^d$, we denote
$\SC^{\sigma}({\bf w},{\bf v},r):=\argmin_{{\bf x}\in\RR^d}\{ \|\sigma({\bf x})-{\bf w}\|^2+r\|{\bf x}-{\bf v}\|^2\},$
where $r>0$ is a proxmial parameter.
If $\sigma({\bf x})={\bf x}$, then $\SC^{\sigma}({\bf w},{\bf v},r)=\frac{{\bf w}+r{\bf v}}{1+r}$; If $\sigma(\cdot)$ is set to be the ReLU function \cite{nair2010rectified},
$[ \SC^{\sigma}({\bf w},{\bf v},r)]_i$ is one of these three items $\{[\frac{{\bf w}+r{\bf v}}{1+r}]_i, [{\bf v}]_i, \min\{{\bf v}_i,0\}\}$ minimizing  $|\sigma(\cdot)-{\bf w}_i|^2+r|\cdot-{\bf v}_i|^2$ \cite{zeng19aICML}.
For loss function with regularization term, we denote an operator as
{\small$\SP^{\LL}({\bf w},{\bf v},r):=\argmin_{{\bf x}\in\RR^d}\{\LL({\bf w}, {\bf x})+\frac{r}{2}\|{\bf x}-{\bf v}\|^2\}.$}
If $\LL({\bf w},{\bf x})$ is set to be $\frac{1}{2}\|{\bf x}-{\bf w}\|^2$, then $\SP^{\LL}({\bf w},{\bf v},r)=\frac{{\bf w}+r{\bf v}}{1+r}$; If $\LL({\bf w},{\bf x})$ is set to be hinge loss $\max(0, 1-w\cdot x)$, then $\SP^{\LL}({\bf w},{\bf v},r)$ is one of these three items $\{v_i, w_i^{-1}, v_i+w_i^{-1} \}$. More proximal operators are widely used in statistical learning community \cite{combettes2011proximal}.

\medskip
\noindent \textbf{Penalty formulation}:
We reformulate the $L$-layer training task as  linearly constrained   optimization problem by introducing $({\bf a}_{l})_{l\in[L-1]}$ and $({\bf W}_{l})_{l\in[L]}$:
\begin{equation}\label{consmodel}
\begin{aligned}
\small
&\min_{{\bf z}, {\bf W}, {\bf a}} \{\LL({\bf y},{\bf z}_{L})+\sum_{l=1}^L R_{l}({\bf W}_l)\},\\
\textrm{s.t.}~~& {\bf z}_{l}={\bf W}_{l}{\bf a}_{l-1},~l\in [L];~ {\bf a}_l=\sigma_l({\bf z}_l), ~l\in [L-1].
\end{aligned}
\end{equation}
If $L=1$, we then need to minimize a function with linear constraints, which can be efficiently solved by ADMM.
When $L>1$, Problem \eqref{consmodel} is a nonlinear constrained problem, which is difficult to solve directly.
Thus, people consider its penalty problem.
Given a very large penalty parameter $\bar{\rho}>0$, we aim to solve a reformulated problem:
\begin{equation}\label{penalty}
\begin{aligned}
\small
&\min_{{\bf z},{\bf W}, {\bf a}} \{\Phi_{\bar{\rho}}({\bf z},{\bf W},{\bf a}):=\LL({\bf y},{\bf z}_{L})+\sum_{l=1}^L R_{l}({\bf W}_l) \\
&\quad+\frac{\bar{\rho}}{2}(\sum_{l=1}^{L}\|{\bf z}_{l}-{\bf W}_{l}{\bf a}_{l-1}\|^2+\sum_{l=1}^{L-1} \|{\bf a}_l-\sigma_l({\bf z}_l)\|^2)\}.
\end{aligned}
\end{equation}
An extreme case is setting $\bar{\rho}=+\infty$, in which \eqref{penalty} is identical to   \eqref{consmodel}.
Actually, even for linearly constrained nonconvex minimization, penalty  method is also a good choice due to that it enjoys much more mild convergent assumption and easily-set parameters than the nonconvex ADMM \cite{sun2019iteratively}.
In the numerical experiments, we update $\bar\rho$ from iteration to iteration, and $\rho_{k+1}$ is set to be $1.1\rho_k$ at the end of each iteration.

It should be noted that the formulation in this work is different from Zeng's work \cite{zeng19aICML}, there is no acitvation $a_L=\sigma_L(z_L)$ before the last layer, and further analysis is based on this formulation.

\medskip
\noindent \textbf{Inertial methods}:
The DLAM is actually the alternating minimization applied to \eqref{penalty}.  In the nonconvex community, the inertial technique \cite{polyak1964some} (also called as heavy-ball or momentum) is widely used and proved to be algorithmically efficient \cite{ochs2014ipiano,pock2016inertial,loizou2017momentum,loizou2018accelerated}. Besides acceleration and good practical performance for nonconvex problems, the advantage of  inertial technique is illustrated by weaker conditions avoiding   saddle points \cite{sun2019heavy}. The procedure of inertial method is quite simple, it uses linear combination of current and last point for next iteration.  For example, the gradient descent minimizing a smooth function $f$ employs the inertial term
${\bf \hat{x}}^{k}={\bf x}^k+\alpha({\bf x}^k-{\bf x}^{k-1})$ ($\alpha\geq 0$)
as
${\bf x}^{k+1}={\bf x}^{k}-\nabla f({\bf \hat{x}}^{k})$.

\medskip
\noindent \textbf{Algorithm}:
We employ the alternating minimization method.
We use the inertial technique for ${\bf z}^k$, ${\bf W}^k$ and ${\bf a}^k$,
\begin{equation*}
\small
\left\{\begin{array}{lll}
\hat{{\bf z}}^k &=& {\bf z}^k+\alpha({\bf z}^k-{\bf z}^{k-1})\\
\hat{{\bf W}}^k &=& {\bf W}^k+\beta({\bf W}^k-{\bf W}^{k-1})\\
\hat{{\bf a}}^k &=& {\bf a}^k+\gamma({\bf a}^k-{\bf a}^{k-1})\\
	{\bf z}_{l}^{k+1} &\in& \displaystyle \argmin_{{\bf z}_l}\{ \Phi_{{\rho}_k}([{\bf z}^k;{\bf z}_l],{\bf W}^k,{\bf a}^k)+\frac{\delta{\rho}_k}{2}\|{\bf z}_l-{\bf \hat{z}}^k_l\|^2\}\\
	{\bf W}_{l}^{k+1} &\in& \displaystyle\argmin_{{\bf W}_l}\{ \Phi_{{\rho}_k}({\bf z}^{k+1},[ {\bf W}^k;{\bf W}_l],{\bf a}^k)+\frac{\delta{\rho}_k}{2}\|{\bf W}_l-\hat{{\bf W}}^k_l\|^2\}\\
	{\bf a}_{l}^{k+1} &\in& \displaystyle \argmin_{{\bf a}_l}\{ \Phi_{{\rho}_k}({\bf  z}^{k+1},{\bf W}^{k+1},[{\bf a}^k;{\bf a}_l])+\frac{\delta{\rho}_k}{2}\|{\bf a}_l-\hat{{\bf a}}^k_l\|^2\}\\
\end{array}
\right..
\end{equation*}
We can see if $\alpha=\beta=\gamma=\delta=0$, the algorithm above then degenerates the DLAM. We first use the linear combinations to predict two new points  $\hat{{\bf W}}^k, \hat{{\bf a}}^k$ (inertial step).
In the substeps of updating ${\bf z}^{k+1}_l$, ${\bf W}^{k+1}_l$ and ${\bf a}^{k+1}_l$, we use a proximal-point way, i.e., adding the regular terms $\frac{\delta{\rho}_k}{2}\|{\bf z}_l-\hat{{\bf z}}^k_l\|^2$, $\frac{\delta{\rho}_k}{2}\|{\bf W}_l-\hat{{\bf W}}^k_l\|^2$ and $\frac{\delta{\rho}_k}{2}\|{\bf a}_l-\hat{{\bf a}}^k_l\|^2$ in the minimizations to get the sufficient descent. Let $d_l$ is the width of the $l$-layer, and we have ${\bf W}_{l}\in \RR^{d_{l-1}\times d_{l}}$.

As we mentioned at the beginning of this section, $\bar{\rho}$ is the penalty parameter, which shall be large to yield the sufficient closed approximate problem.
A natural problem is that large  $\bar{\rho}$ leads to small change of $({\bf z}_{L}^{k})_{k\geq 1}$ and $({\bf W}^{k})_{k\geq 1}$, which indicates   the algorithm  slowed down in this case.
To overcome this drawback, we set a small $\rho_0>0$ as the initialization and then increase it in the iterations as $\rho_{k+1}=\min\{\theta\rho_{k},\bar{\rho}\}$, where $\theta>1$.
Such techniques   have been used in the image processing and nonconvex linearly constrained problems \cite{wang2008new,sun2019bregman}.
With increasing penalty parameter strategy,  our algorithm can be  displayed in Algorithm \ref{alg1}.
\begin{algorithm}
\small
\caption{Inertial Proximal Deep Learning Alternating Minimization for Neutral Network Training (iPDLAM)}
\begin{algorithmic}\label{alg1}
\REQUIRE   parameters  $\alpha\geq 0$, $\beta\geq 0$, $\gamma>0$, $\delta>0$, $\theta>1$, $\bar{\rho}\geq\rho_0>0$, $K>0$\\
\textbf{Initialization}: $({\bf z}_{l}^0)_{l\in[L]}$, $({\bf a}_{l}^0)_{l\in[L-1]}$, $({\bf W}_{l}^0)_{l\in[L]}$\\
\textbf{for}~$k=1, 2, \ldots$, $K$ \\
~~~1. $\hat{{\bf z}}^k={\bf z}^k+\alpha({\bf z}^k-{\bf z}^{k-1})$;\\
~~~2. $\hat{{\bf W}}^k={\bf W}^k+\beta({\bf W}^k-{\bf W}^{k-1})$;\\
~~~3. $\hat{{\bf a}}^k={\bf a}^k+\gamma({\bf a}^k-{\bf a}^{k-1})$;\\
~~~4. ${\bf z}_{l}^{k+1}=\SC^{\sigma_l}({\bf a}_l^k, \frac{{\bf W}_{l}^k{\bf a}_{l-1}^k+\delta{\bf \hat{z}}^k_l}{1+\delta},1+\delta),~i\in[L-1]$;\\
~~~5. ${\bf z}_{L}^{k+1}=\SP^{\LL}({\bf y},\frac{{\bf W}_{L}^k{\bf a}_{L-1}^k+\delta\hat{{\bf z}}^k_L}{1+\delta}, \rho_k(1+\delta))$;\\
~~~6. Update ${\bf W}_{l}^{k+1}$;\\
~~~7. Update ${\bf a}_{l}^{k+1}$;\\
~~~8. Update $\rho_{k+1}=\min\{\theta\rho_{k},\bar{\rho}\}$\\
\textbf{end for}
\end{algorithmic}
\end{algorithm}

\section{Numerical Experiments}
In this section, we present the numerical results of our algorithm.
We follow the experimental setup introduced by \cite{wang2019admm}.
Specifically, we consider the DNN training model \eqref{consmodel} with ReLU activation, the squared loss, and the network architecture being an MLPs with hidden layers, on the two datasets, MNIST \cite{lecun1998gradient} and Fashion MNIST \cite{xiao2017fashion}.
The specific settings were summarized as follows:
\begin{itemize}
\item For the MNIST data set, we implemented a 784-(1500$\times$2)-10 MLPs (\textit{i.e.}, the input dimension d0 = 28 $\times$ 28 = 784, the output dimension d3 = 10, and the numbers of hidden units are all 1500), and set $\alpha = \beta = \gamma, \delta$ with different values to testify the proposed algorithm iPDLAM. The sizes of training and test samples are 60000 and 10000, respectively.
\item The parameters of the DLAM is adopted as the default parameters given in \cite{zeng19aICML}. The learning rate of SGD and its variants ({\textit .i.e.} RMSprop, Adam, Adadelta, AMSGrad, Adamax.) is 0.001 (a very conservative learning rate to see if SGD can train the DNNs). More greedy learning rates such as 0.01 and 0.05 have also been used, and similar results of training are observed.
\item For each experiment, we used the same mini-batch sizes (512) and initializations for all algorithms. Specifically, all the weights ${\bf W}_l^0$ are randomly initialized from a Gaussian distribution with a standard deviation of 0.01 and the bias vectors are initialized as vectors of all 0.1, while the auxiliary variables ${\bf z}_l$ and state variables ${\bf a}_l$ are initialized by a single forward pass.
\end{itemize}


The proposed algorithm is implemented based on PyTorch.
And the experiments are conducted on a desktop computer with Intel\textsuperscript{\textregistered} Core\textsuperscript{\texttrademark} CPU i7-8700k @ 4.70 GHz 
, 16 GB memory, and running Ubuntu 18.04 Server OS. 
The CPU contains 12 physical cores. We use PyTorch 1.5.0 to implement the proposed algorithm.

\subsection{The closed solutions of substeps in iPDLAM}
Given the activation $\sigma_l(\cdot)$ to be ReLU function, the loss function is $\ell_2$-norm, $R_l(\cdot)$ is 0, and the bias parameter added to the network, the closed solution of substeps in iPDLAM is presented in this subsection. The parameters are discussed in the following subsections. Besides the proximal operators on ${\bf z}_l^{k+1}$ for $l\in[L]$, the closed solutions for ${\bf W}_{l}^{k+1}$ and ${\bf b}_{l}^{k+1}$  for $l\in[L]$ are:
\begin{equation*}
\small
{\bf W}_{l}^{k+1} = (({\bf z}_l^{k+1}-{\bf b}_l^k)({\bf a}_{l-1}^k)^T + \delta{\bf \hat w}_l^k)(({\bf a}_{l-1}^k) ({\bf a}_{l-1}^k)^T + \delta {\bf I})^{-1},\]
\[{\bf b}_{l}^{k+1} = ({\bf z}_l^{k+1}- {\bf W}_l^{k+1} {\bf a}_{l-1}^k + \delta*\hat{{\bf b}}^k_l)/(1+\delta),
\end{equation*}
in which, ${\bf a}_0$ is the input data.

In the following, we present numerical experiments to verify the theoretical analysis, the iPDLAM's superem performance over baselines under different $\alpha, \beta, \gamma$ and $\delta$. And further more, we discuss the variables update order's impect on the final performance.

\subsection{Comparisons with DLAM}

\begin{figure}
\centering
\includegraphics[width=0.2\textwidth]{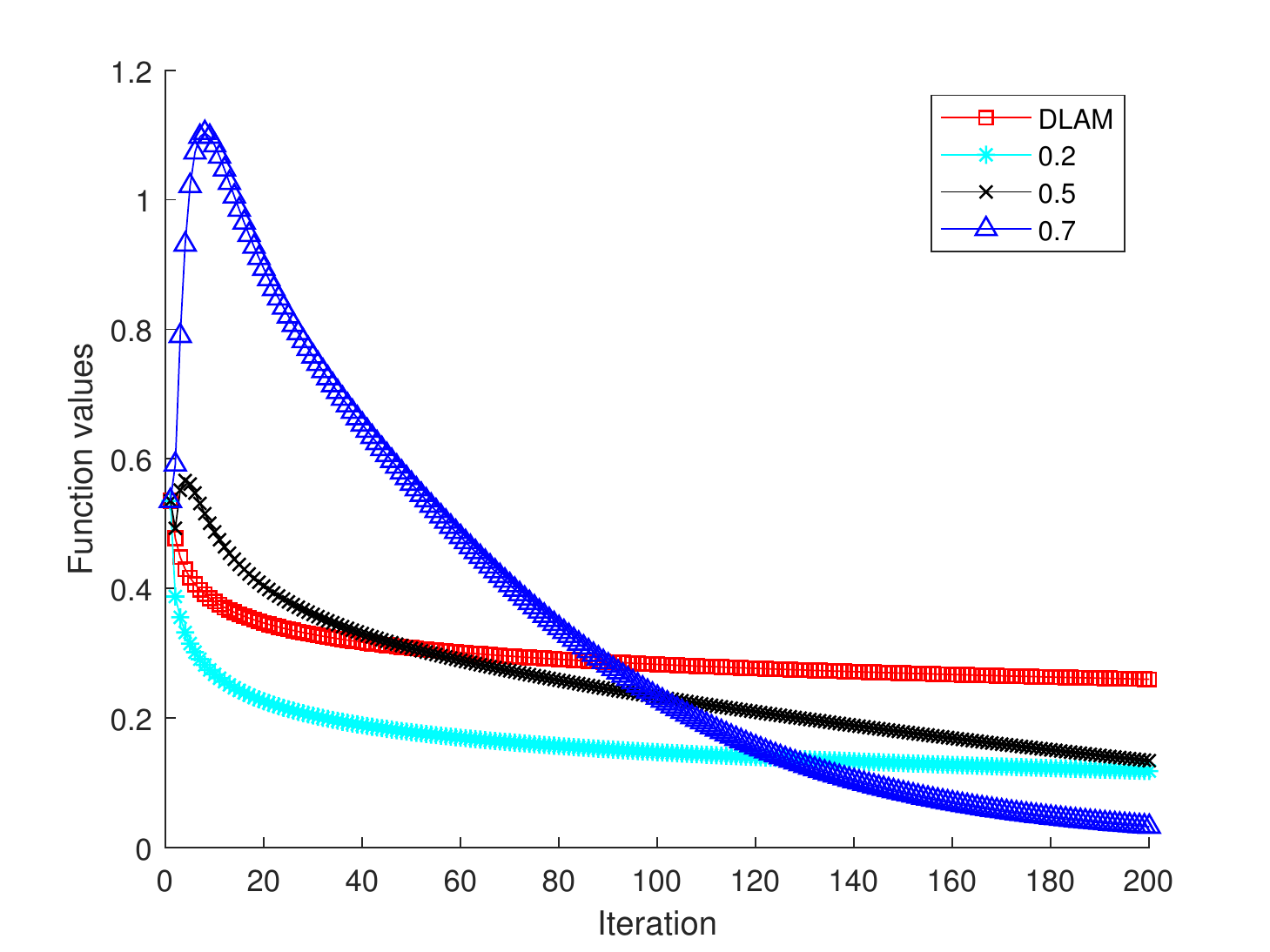}
\includegraphics[width=0.2\textwidth]{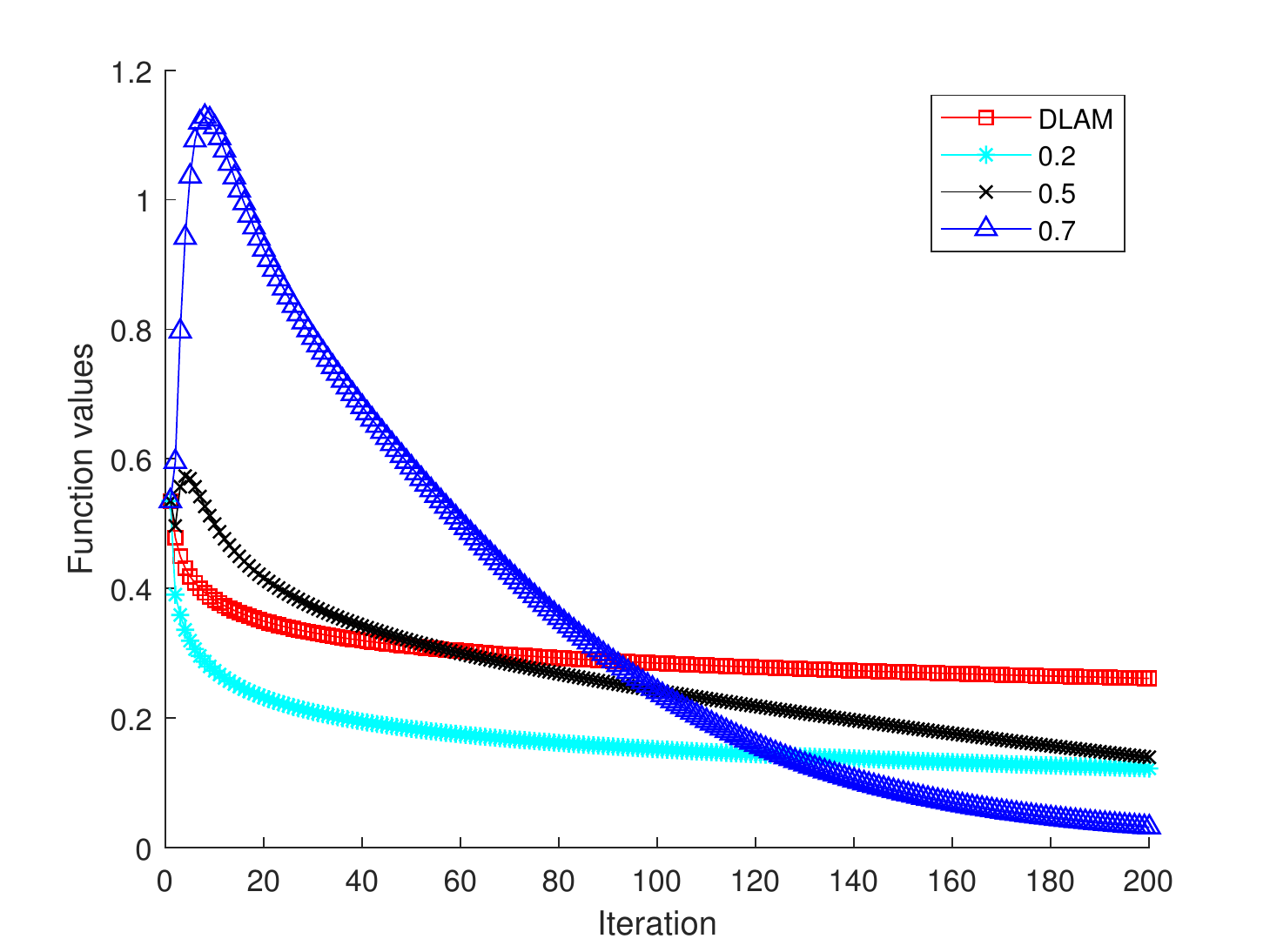}
\includegraphics[width=0.2\textwidth]{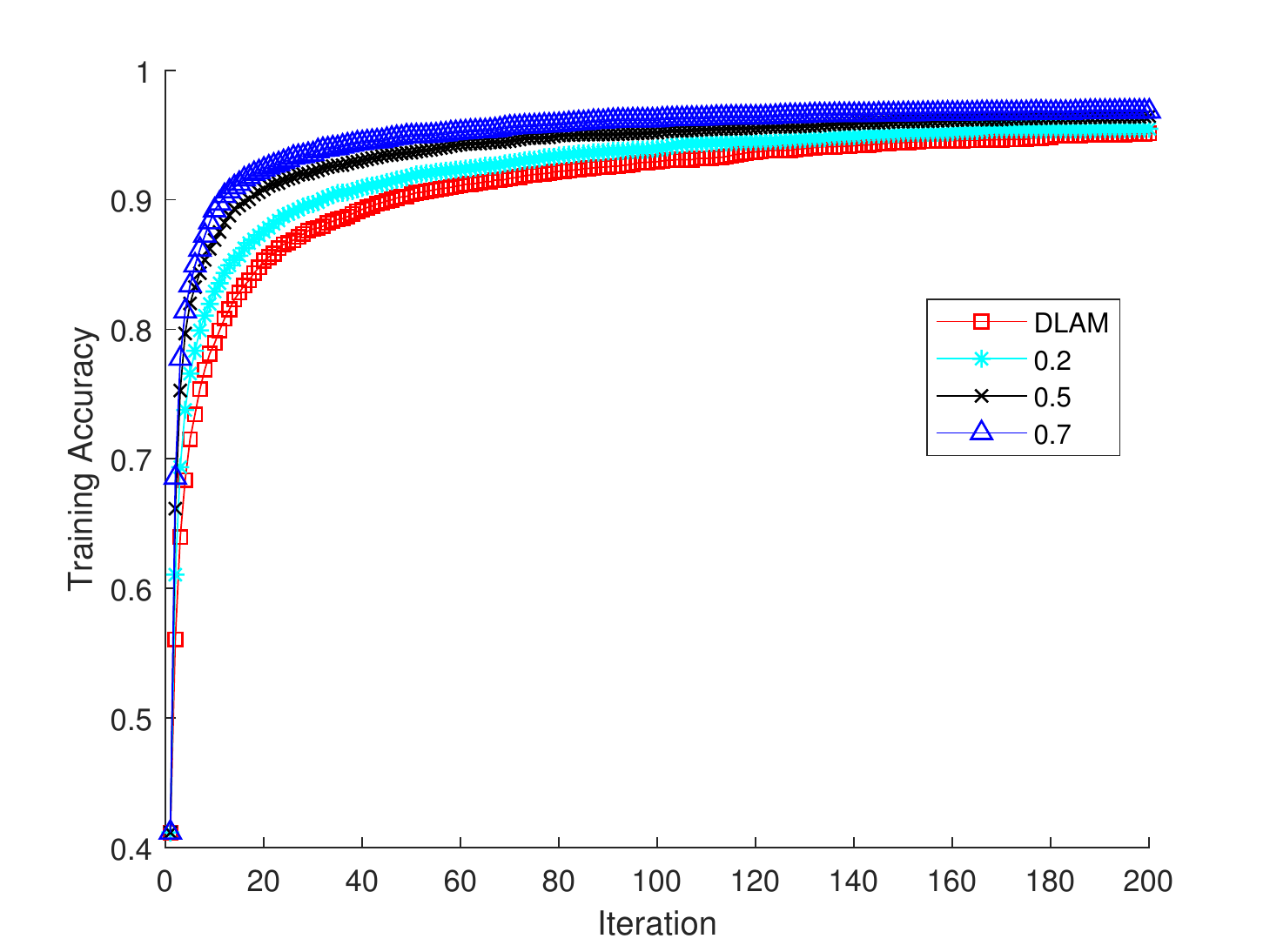}
\includegraphics[width=0.2\textwidth]{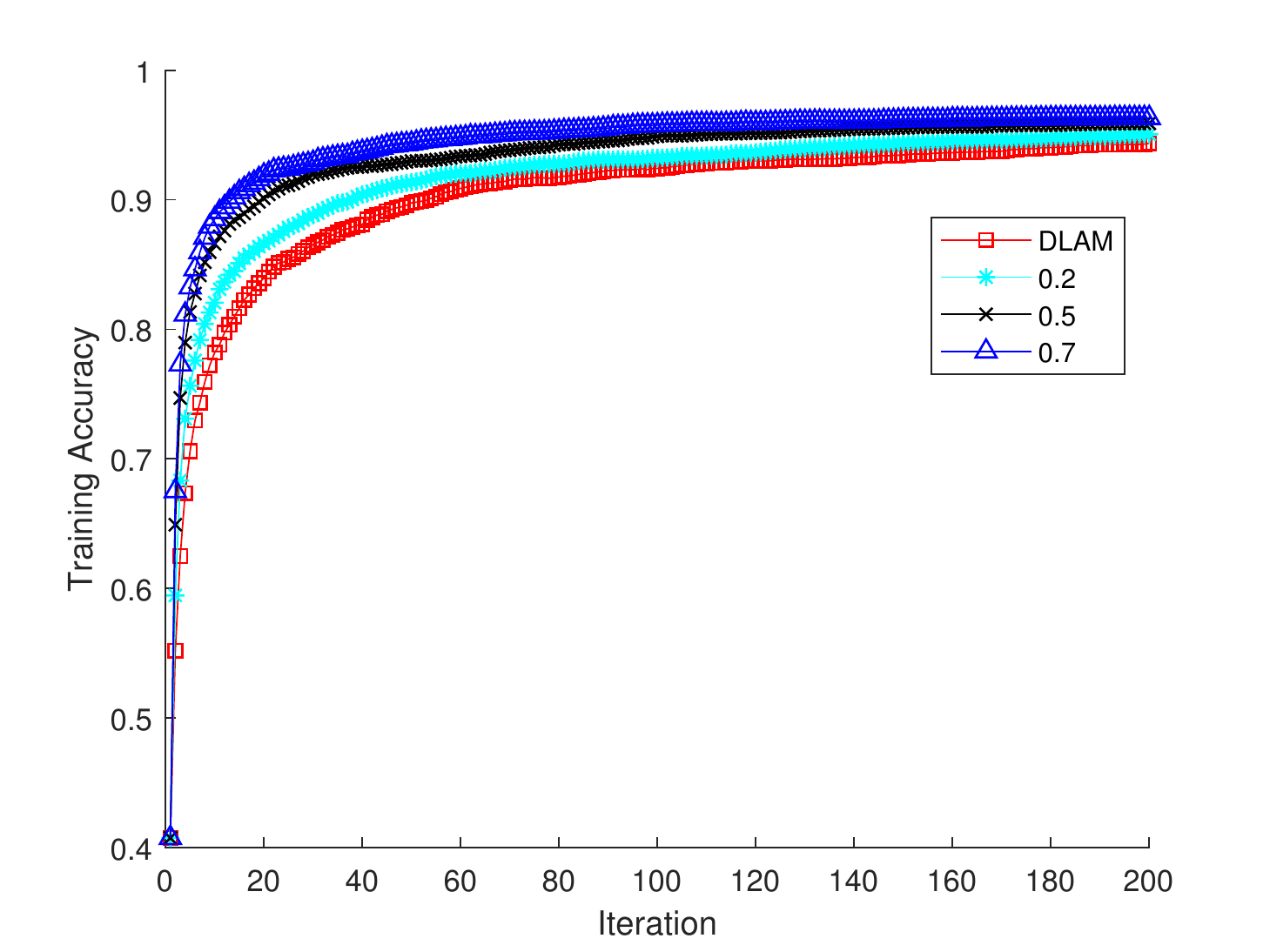}
\includegraphics[width=0.2\textwidth]{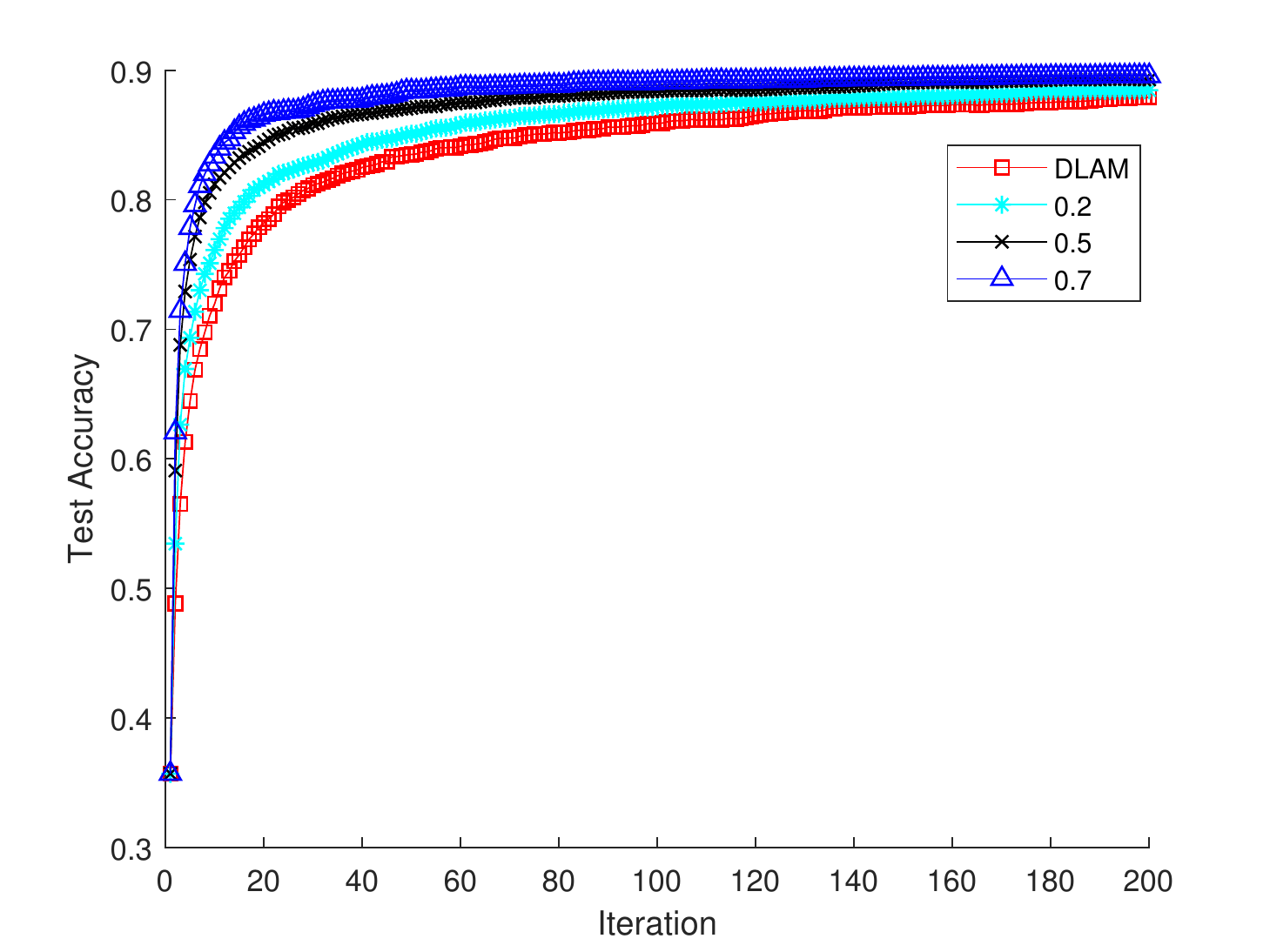}
\includegraphics[width=0.2\textwidth]{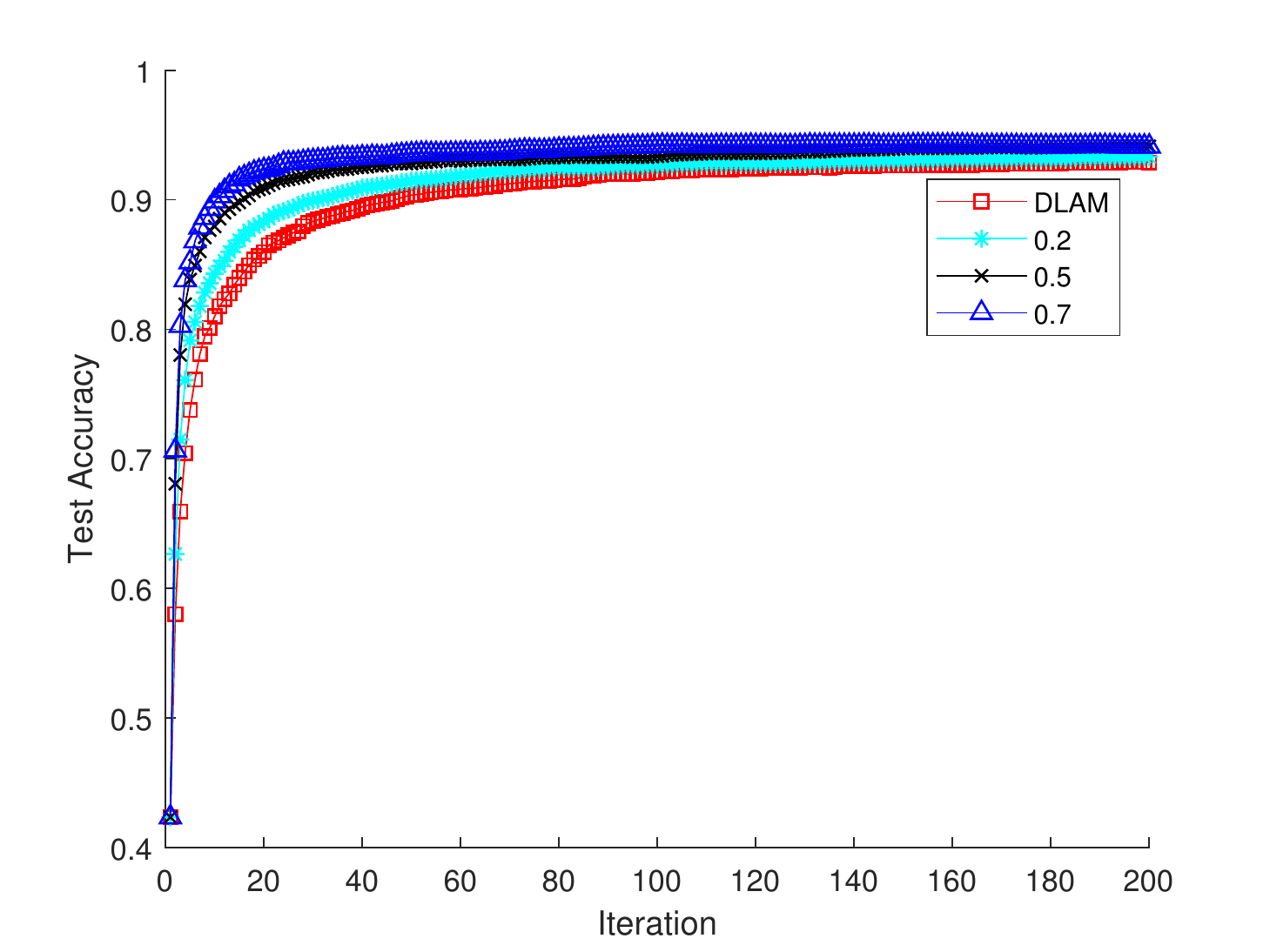}
\caption{Comparations with DLAM on Function values, training and test accuracy versus the iteration when $\alpha=\beta=\gamma=0.2,0.5,0.7$ on MNIST (left) and Fashion MNIST (right).}
\label{fig1}
\end{figure}

Firstly, we present the objective function values, training accuracy and test  accuracy when $\alpha=\beta=\gamma=0.2,0.5,0.7$ for the two datasets; $\delta$ is set as $1$. The maximum iteration is set as $200$. The results of test 1 are given in Figure \ref{fig1}.
 We can see that iPDLAM outperforms  DLAM: iPDLAM can achieve smaller objective function values and larger training and test accuracy.  It is worth mentioning that the curves of the function values are not always descent. This fact does not contradict our theory due to that we prove $(\xi_k)_{k}$ is decreasing after sufficient iteration (Lemma 1) rather than the  function values.

From Table \ref{tab1}, we can see that the inertial term contributes when $0<\alpha<1$. When $\alpha>1$, iPDLAM performs poorly. This phenomenon coincides with our theoretical findings.

\begin{table}[htb!]
\centering
\scriptsize
\begin{tabular}{|r|rrr|rrr|}
\cline{2-7}
\multicolumn{1}{c|}{} & \multicolumn{3}{c|}{MNIST}
       & \multicolumn{3}{c|}{Fashion MNIST} \\
\hline
\rowcolor[gray]{0.9}
$\alpha$ & FV & TrA & TeA & FV & TrA & TeA   \\
0.1 &    0.14 &     95.4\% & 88.2\% & 0.14 &     94.6\% &    93.2\%   \\
\rowcolor[gray]{0.9}
0.2 &    0.12 &    95.7\% & 88.9\% & 0.12  &    95.0\% &     93.4\%    \\
0.3 &    0.13 &     95.8\% & 89.2\% & 0.13 &    95.5\% &     93.7\%   \\
\rowcolor[gray]{0.9}
0.4 &    0.14 &      96.2\% &89.5\% & 0.14 &    95.7\% &      93.9\%  \\
0.5&    0.14 &      96.4\% & 89.3\% & 0.14 &    96.2\% &      94.1\%  \\
\rowcolor[gray]{0.9}
0.6&    0.10&       96.6\%& 89.2\% & 0.10 &    96.6\% &     94.2\%   \\
0.7&    0.03 &      96.8\% & 89.6\%  & 0.09 &    96.3\% &      94.2\%   \\
\rowcolor[gray]{0.9}
0.8&    0.28 &      92.0\% & 85.8\% & 0.29 &    91.6\% &     91.8\%   \\
0.9&    0.61 &      95.1\% & 87.6\% & 0.64 &   94.6\% &      93.0\%   \\
\rowcolor[gray]{0.9}
1.0&    5510 &      89.0\% & 78.1\% & 5600 &    89.1\% &       86.2\%   \\
1.1&    nan &      nan & nan & nan &    10.4\% &      10.4\%   \\
\rowcolor[gray]{0.9}
1.2&    nan &      nan & nan & nan &    10.4\% &     10.4\%   \\
\hline
\end{tabular}
\caption{Function values, training accuracy and test  accuracy for different $\alpha$ ($=\beta=\gamma$)   on MNIST and Fashion MNIST after  200 iterations.   \emph{Function Values} (\texttt{FV}),  \emph{Training Accuracy} (\texttt{TrA}) and \emph{Test Accuracy} (\texttt{TeA}).  \label{tab1}}
\vspace{-2em}
\end{table}

\subsection{Robust performance on different values of $\delta$}
\begin{figure}[h]
\centering
\includegraphics[width=0.2\textwidth]{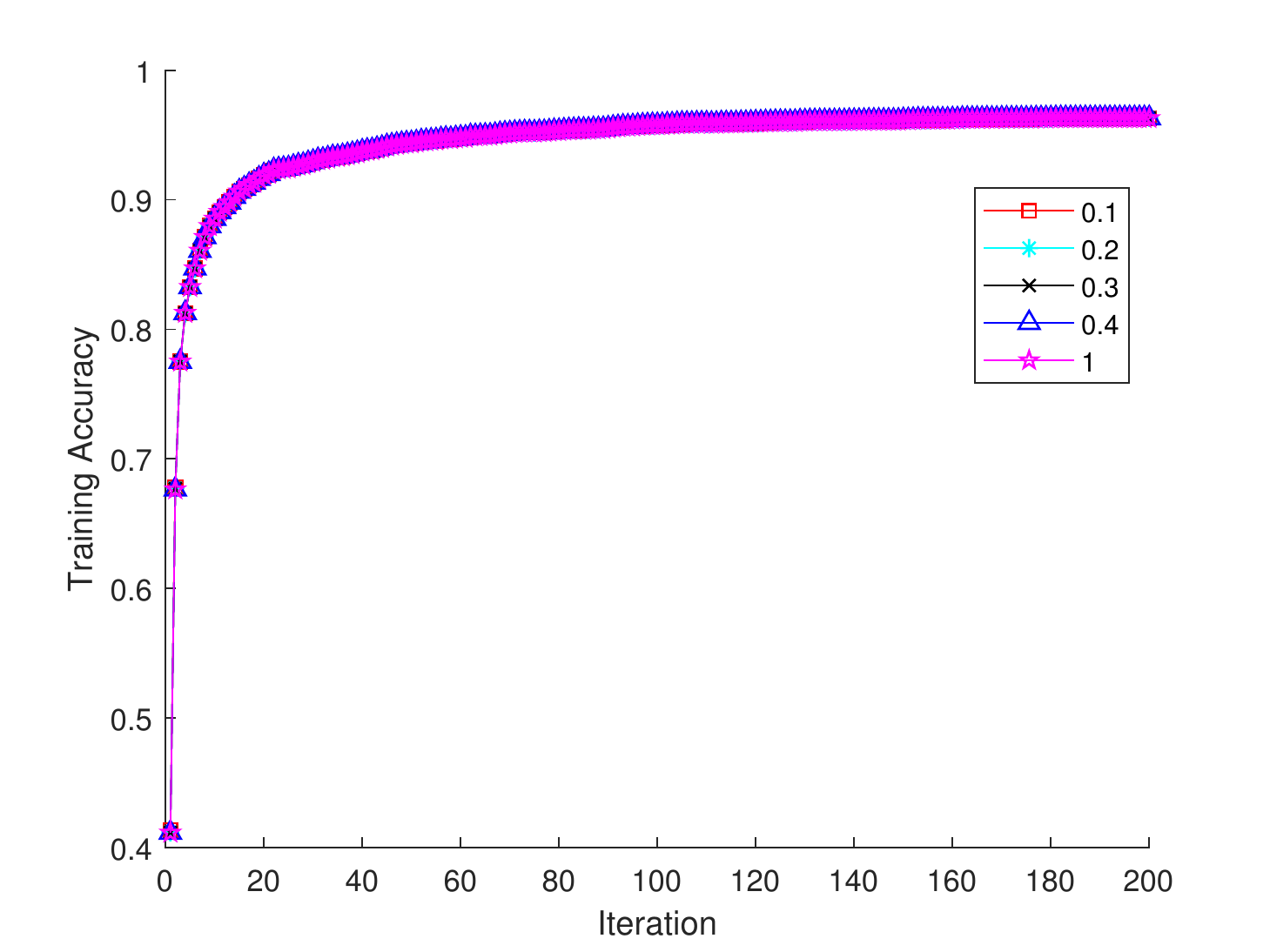}
\includegraphics[width=0.2\textwidth]{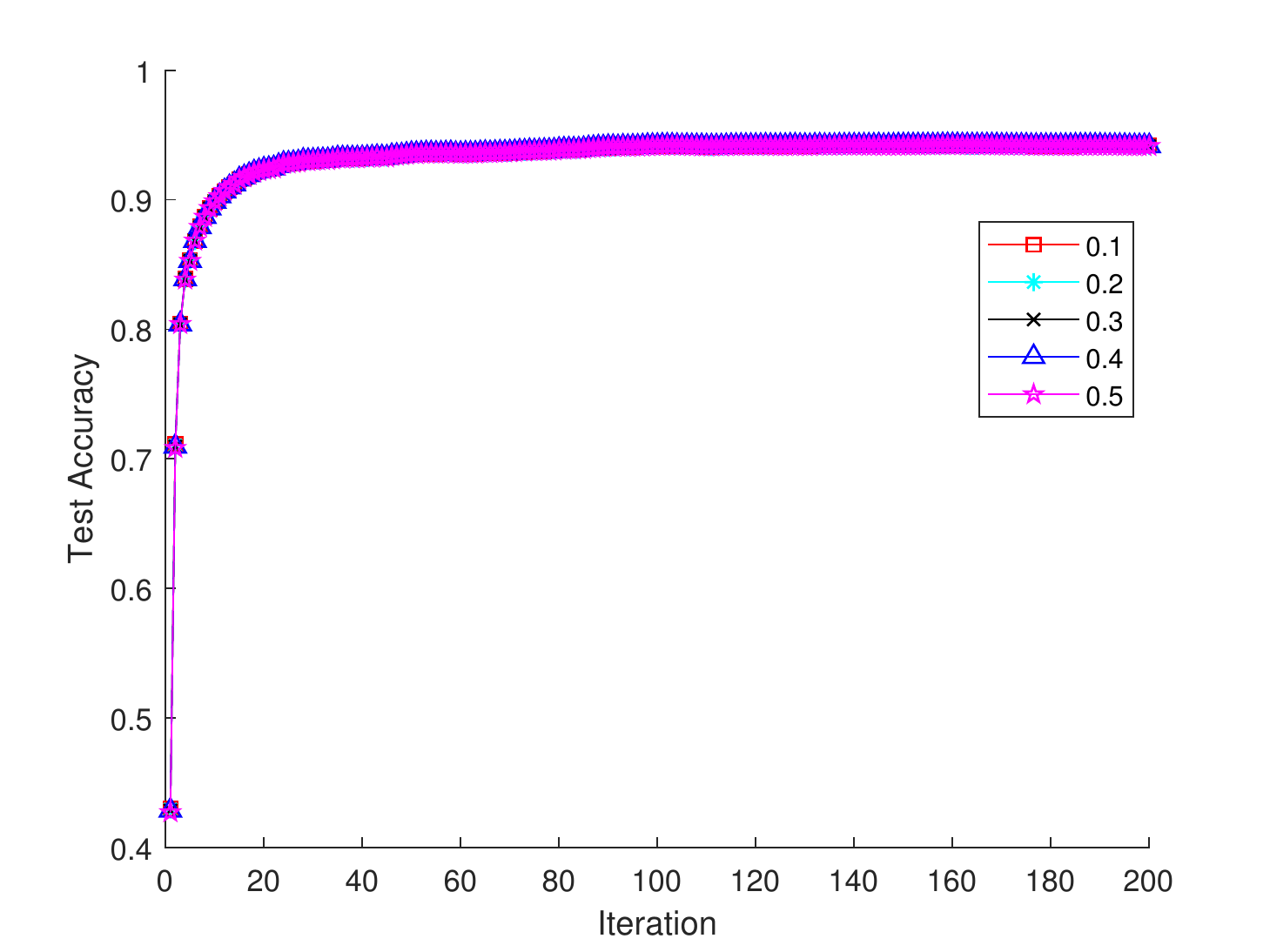}
\caption{Training  and test  accuracy versus the iteration for different $\delta$.}
\label{fig2}
\end{figure}

In the second test, we use $\alpha=\beta=\gamma=0.7$, the parameter $\delta$ is set as $0.1,0.2,0.3,0.4,1$. The dataset is the MNIST. The results of the second test are given in Figure \ref{fig2}. In the five cases, the training and test accuracy versus the iterations actually perform very similar for the five cases. The results show that the algorithm is insensitive  to  $\delta$.

\subsection{Results against classical Deep Learning optimizers}
\begin{figure}[h]
\centering
\includegraphics[width=0.2\textwidth]{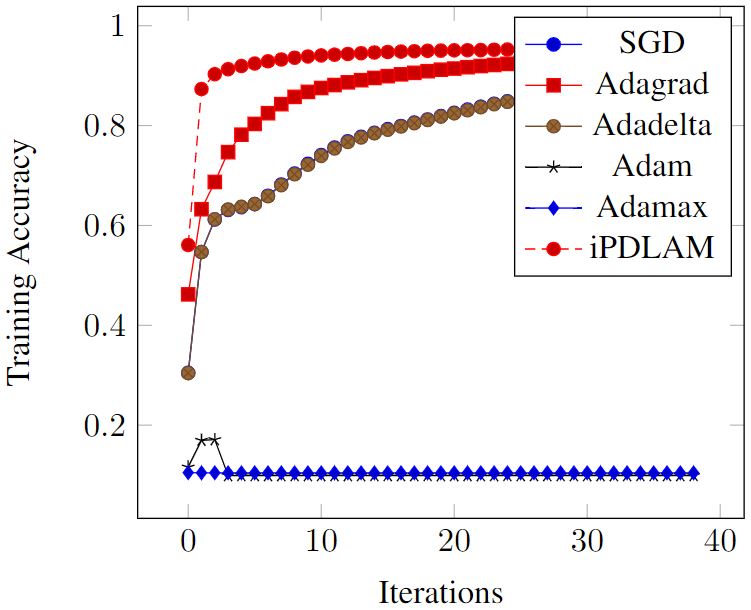}
\includegraphics[width=0.2\textwidth]{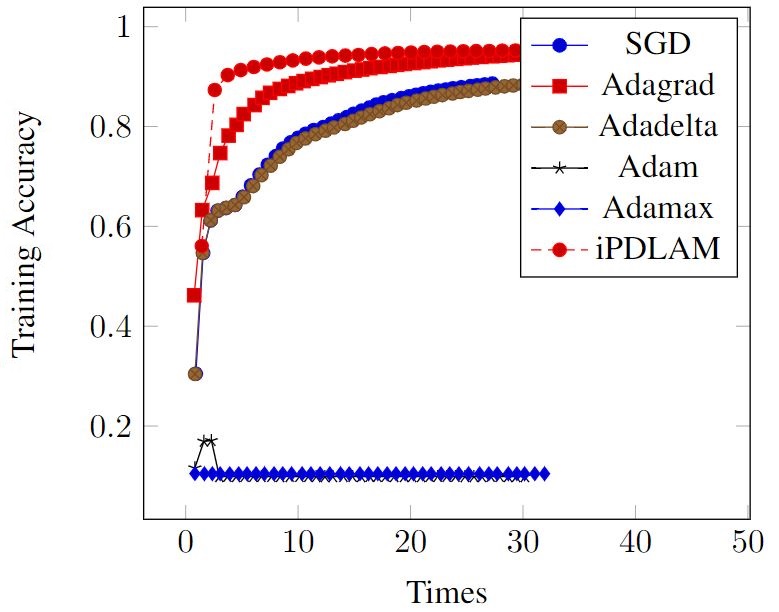}
\vspace{-1em}
\caption{Training accuracy versus the iteration and time for different algorithms on MNIST.}
\label{fig3}
\vspace{-1.5em}
\end{figure}

We compare iPDLAM with SGD, AdaGrad\cite{duchi2011adaptive}, Adadelta\cite{ADADELTA}, Adam and Adamax\cite{kingma2014adam}. The training and test accuracy versus the iteration (epoch) for different algorithms are reported in Figure \ref{fig3}. 
Although our algorithm cannot beat classical algorithms on training accuracy, it performs better than most of them on test accuracy.
Our algorithm can learn a quite good parameter in very small iterations (less than 10). 
Moreover, the proposed iPDLAM is alawys better than classical optimizers at first 40 iterations, both in  It is also the fastest one to reach stable, even when SGD fails to converge (the blue one on the bottom in Figure \ref{fig3}).

Based on the experiment results, it is recomended to adopt the iPDLAM to train the neural networks at the first stage as a warm-up strategy, which is expected to greatly reduce the total training time cost, and speed up the training process.

\subsection{The update order of variables}
\begin{figure}[h]
\centering
\includegraphics[height=3cm, width=6cm]{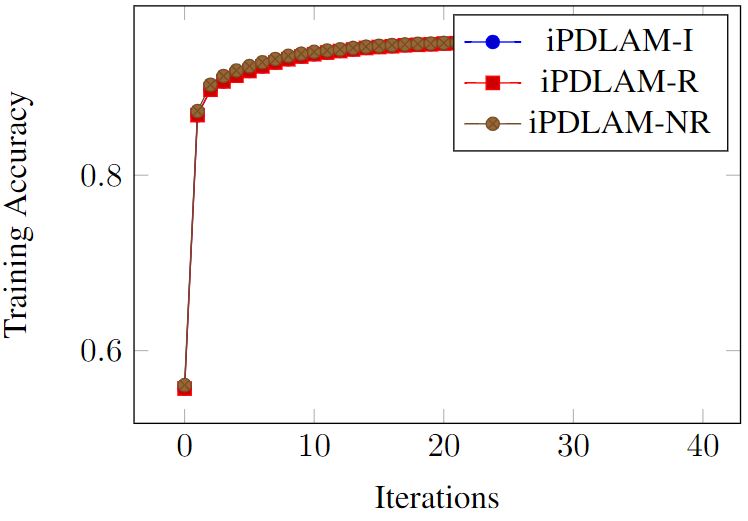}
\vspace{-1em}
\caption{The variables update order has a little impact on the optimization process.}
\label{fig4}
\vspace{-2.5em}
\end{figure}

We compare the alternating minimization methods and use the following cyclic orders to update the variables:
\begin{itemize}
\small
\item {\textbf{Reverse:}} ${\bf z}_L \rightarrow {\bf z}_{L-1} \rightarrow \ldots \rightarrow {\bf z}_1  \rightarrow {\bf W}_L \rightarrow {\bf W}_{L-1} \rightarrow\ldots \rightarrow {\bf W}_1 \rightarrow {\bf a}_L  \rightarrow  {\bf a}_{L-1} \rightarrow \ldots \rightarrow {\bf a}_1 \rightarrow {\bf z}_L \ldots$
\item {\textbf{Increase:}} ${\bf z}_1 \rightarrow {\bf z}_2 \rightarrow \ldots \rightarrow {\bf z}_L  \rightarrow {\bf W}_1 \rightarrow {\bf W}_2 \rightarrow\ldots \rightarrow  {\bf W}_L \rightarrow {\bf a}_1  \rightarrow  {\bf a}_2 \rightarrow \ldots \rightarrow {\bf a}_{L-1} \rightarrow {\bf z}_{1}  \ldots$
\item {\textbf{Nested Reverse (NR):}} ${\bf z}_L \rightarrow {\bf W}_L \rightarrow {\bf a}_{L-1}\rightarrow {\bf z}_{L-1} \rightarrow {\bf W}_{L-1} \rightarrow\ldots \rightarrow {\bf a}_2\rightarrow {\bf z}_2 \rightarrow {\bf W}_2 \rightarrow {\bf a}_1\rightarrow {\bf z}_1 \rightarrow {\bf W}_1  \rightarrow {\bf z}_L \rightarrow \ldots$
\end{itemize}
The numerical results are reported in Figure~\ref{fig4}.
The variables update order has little impact on the optimization process, which suggests iPDLAMs could be further applied to parallel and distributed settings. Further experiments results are provided in the supplymentary material, and the implementation of this work is public available.

\section{Conclusion}
In this paper, we propose an improved  alternating minimization, named \textit{iPDLAM}, for the neural network training. The development of the algorithm is based on the inertial techniques applied to the penalty formulation of the training task. Different from the stochastic training methods, our algorithm enjoys solid convergence guarantee, and the numerical results show that the  proposed algorithm takes smaller iterations to reach the same training and test accuracy compared with various classical training algorithms.

\small
\bibliographystyle{ieeetr}
\bibliography{inerbib}

\end{document}